\documentclass[10pt,twocolumn,letterpaper]{article}

\usepackage[pagenumbers]{cvpr} %

\definecolor{cvprblue}{rgb}{0.21,0.49,0.74}
\usepackage[pagebackref,breaklinks,colorlinks,allcolors=cvprblue]{hyperref}
\usepackage{multirow}

\title{[CLS] Token Tells Everything Needed for Training-free Efficient MLLMs}

\author{Ao Wang$^{1}$\thanks{Equal contribution.} \quad Fengyuan Sun$^{1}$\footnotemark[1] \quad Hui Chen$^{2}$ \quad Zijia Lin$^{1}$ \quad Jungong Han$^{3}$ \quad Guiguang Ding$^{1}$ \\
		$^1$School of Software, Tsinghua University \quad $^2$BNRist, Tsinghua University \\ $^3$Department of Automation, Tsinghua University\\}

\begin{document}
\maketitle

\begin{abstract}
Multimodal Large Language Models (MLLMs) have recently demonstrated strong performance across a wide range of vision-language tasks, garnering significant attention in the computer vision. However, their efficient deployment remains a substantial challenge due to high computational costs and memory requirements. Recognizing the redundancy of information within the vision modality, recent studies have explored methods for compressing visual tokens in MLLMs to enhance efficiency in a training-free manner. Despite their effectiveness, existing methods like FastV~\cite{chen2024image} rely on the attention between visual tokens and prompt text tokens as the importance indicator, overlooking the relevance to response text and thus introducing perception bias. In this paper, we demonstrate that in MLLMs, the [CLS] token in the visual encoder inherently knows which visual tokens are important for MLLMs. Building on this prior, we introduce a simple yet effective method for train-free visual token compression, called VTC-CLS. Firstly, it leverages the attention score of the [CLS] token on visual tokens as an importance indicator for pruning visual tokens. Besides, we also explore ensembling the importance scores derived by the [CLS] token from different layers to capture the key visual information more comprehensively. Extensive experiments demonstrate that our VTC-CLS achieves the state-of-the-art performance across various tasks compared with baseline methods. It also brings notably less computational costs in a training-free manner, highlighting its effectiveness and superiority. Code and models are available at \url{https://github.com/THU-MIG/VTC-CLS}.
\end{abstract}

\section{Introduction}
Recent years have seen a significant revolution of Multimodal Large Language Models (MLLMs)~\cite{liu2024visual,zhu2023minigpt,liu2024improved,bai2023qwenllm,wang2023cogvlm}. Benefiting from notable advancements of the Large Language Models (LLMs)~\cite{touvron2023llama,touvron2023llama2,dubey2024llama,team2023gemini,bai2023qwen,glm2024chatglm}, MLLMs have shown the powerful understanding and generation capabilities for multimodal information, exhibiting immense application potential in various areas~\cite{niu2024llarva,li2024llava,gu2024anomalygpt}. They usually incorporate pretrained visual encoder and connector to transform the input image into sequential visual tokens which are then fed into a LLM~\cite{liu2024visual,liu2024improved,li2023blip,zhu2023minigpt,wang2023cogvlm}. Through the instruction tuning on high-quality datasets, the visual information and textual semantics are well aligned, achieving effective multimodal capabilities.

\begin{figure}[t]
    \centering
    \includegraphics[width=0.9\linewidth]{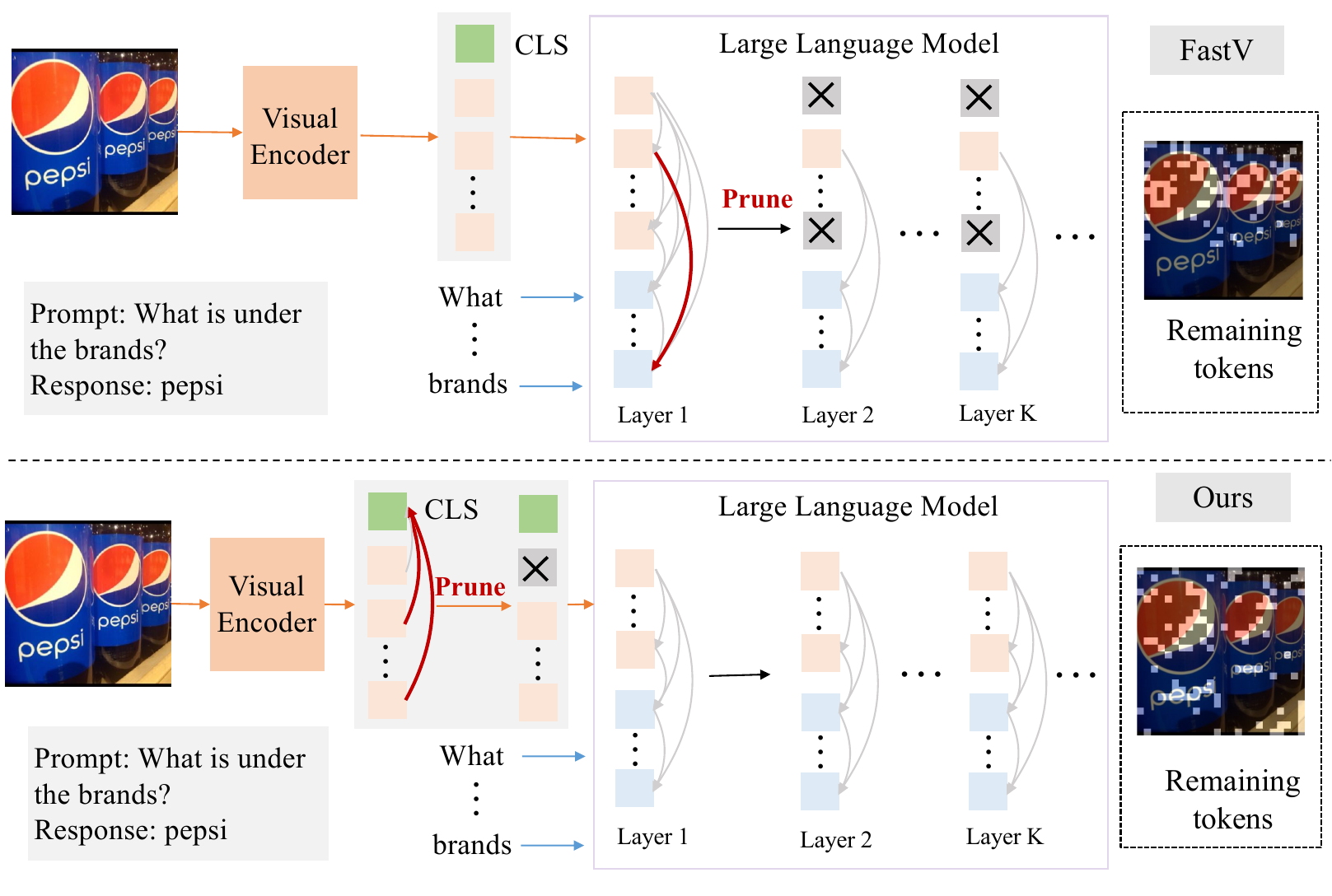}
    \caption{Top: the framework of FastV. It uses the attention in the LLM to perform token pruning, suffering from a perception bias that tokens that are relevant to the response text are inadvertently eliminated. Bottom: the framework of our VTC-CLS. We use [CLS] token to identify salient visual tokens, maintaining comprehensive visual information for response generation. Red lines means strong correlations. Orange and blue rectangles denotes the visual token and the text token, respectively.}
    \label{fig:intro}
    \vspace{-15pt}
\end{figure}

Despite the inspiring performance, the introduction of visual signals for LLMs also brings significant computational complexity and memory consumption due to the large number of visual tokens, increasing the inference overhead notably. For example, LLaVA-1.5~\cite{liu2024improved} transforms 336$\times$336 and 672$\times$672 images into 576 and 2304 visual tokens, respectively. Recognizing this, some previous works explore designing compact connectors. For example, MobileVLM variants~\cite{chu2023mobilevlm,chu2024mobilevlm} introduce the lightweight downsample projector to reduce 75\% visual tokens. However, such ways necessitate substantial resources for designing and training, limiting its application in practice. 

Recently, FastV~\cite{chen2024image} observed that directly removing less important visual tokens based on the attention from the LLM does not significantly harm the performance of MLLMs. This characteristic is appealing because it allows for model pruning without the need for an additional, costly training process, thereby introducing a new paradigm of \texttt{training-free} pruning for MLLMs. However, despite its effectiveness, we identify a critical issue with FastV related to perception bias during visual token pruning. Specifically, FastV uses attention scores between visual tokens and prompt text tokens as the importance indicator (Top in \cref{fig:intro}). This approach overlooks the relevance between the input image and the text to be generated. As a result, the remaining visual tokens are more closely related to the prompt text than to the intended generated response. Consequently, during generation, FastV may discard crucial visual context that would benefit the response, leading to perception bias.

To address this issue and ensure comprehensive visual perception during pruning, we first recognize that the visual encoder inherently contains valuable prior knowledge about which visual tokens are important for MLLMs.  Typically, a pretrained CLIP-ViT~\cite{radford2021learning} is used as the visual encoder in MLLMs. Previous works reveal that  the [CLS] token in CLIP image encoder captures key visual information for the image~\cite{liang2022not}. Consequently, the association with the [CLS] token can indicate significant token with key visual signals of the image. Intuitively, in the attention mechanism of LLMs, visual tokens with key information will be more frequently attended to by textual tokens, facilitating accurate visual perception. This implies that important tokens in the visual encoder are more likely to align with those required by the LLM for both the prompt text and the generated response. We thoroughly analyze this hypothesis in \Cref{sec:consistency} by comparing important tokens identified by the [CLS] token in the visual encoder with those highlighted by the attention mechanism in LLMs. Our analysis reveals that \textit{there is a high level of consistency between the important tokens in the visual encoder and those in the LLM}. This suggests that the [CLS] token actually tells all the relevant visual information needed by the LLM, allowing for efficient reduction of visual tokens based on their relationship with the [CLS] token before the process of LLM.

Based on the above observation, we propose to leverage the prior knowledge about the association of the [CLS] token with visual tokens in the visual encoder to effectively evaluate the importance of visual tokens and perform Visual Token Compression to shorten the visual context, dubbed VTC-CLS. Specifically, we utilize the attention score distribution of [CLS] token to identify significant tokens with key visual information. Leveraging the observed consistency between token importance in the visual encoder and the LLM, we can preserve comprehensive visual information for LLMs after pruning. Besides, we further propose to ensemble the importance distributions across layers and utilize the joint selection for critical visual tokens. This further strength the consistency between important visual tokens in the visual encoder and LLM, thereby bringing performance improvements.
Our VTC-CLS is simple and can serve as a plug-and-play method to accelerate the inference of MLLMs in a training free manner, showing high practicality. We conduct extensive experiments to assess its effectiveness and efficiency across various visual-language tasks~\cite{lu2022learn,singh2019towards,liu2023mmbench,li2023evaluating,yu2023mm,hudson2019gqa,li2024seed}. It demonstrates the state-of-the-art performance compared with existing works. Besides, It also provides notable inference acceleration for MLLMs, highlighting the efficiency.

In summary, our contributions are three-folds:
\begin{enumerate}
    \item We reveal the strong consistency of important visual tokens in the visual encoder and LLM, thus advocating for a novel perspective on visual token compression from the abundant prior knowledge in the visual encoder.
    \item We propose VTC-CLS for compressing visual signals for MLLMs. It leverages the attention of the [CLS] token on visual tokens to reserve important ones needed by LLMs. Moreover, it ensembles the selection across layers for comprehensive compression.
    \item We conduct extensive experiments to demonstrate the superiority of our method over others. It exhibits the significant performance improvements compared with existing works while also accelerating inference notably.
\end{enumerate}

\section{Related Work}
\textbf{Multimodal Large Language Models.}
In recent years, Large Language Models (LLMs)~\cite{touvron2023llama,touvron2023llama2,dubey2024llama,team2023gemini,bai2023qwen,glm2024chatglm} have exhibited remarkable language understanding and generation capabilities through pretrained on large-scale textual corpus. Inspired by this, numerous works have explored to improve their multimodal perception and comprehension by incorporating the visual encoders, delivering various Multimodal Large Language Models (MLLMs)~\cite{liu2024visual,zhu2023minigpt,liu2024improved,chen2020imram,chen2018show,bai2023qwenllm,wang2023cogvlm}. For example, Flamingo~\cite{alayrac2022flamingo} and BLIP-2~\cite{li2023blip} bridge the modality gap between pretrained LLM and visual encoders, showing the emerging ability of performing zero-shot visual-language tasks. MiniGPT-4~\cite{zhu2023minigpt} and LLaVA~\cite{liu2024visual} introduce instruction tuning on high-quality visual-language datasets, enhancing the multimodal generation ability of MLLMs for complex visual-language tasks. 

Motivated by the impressive performance of MLLMs, like works on efficient vision models~\cite{howard2017mobilenets,radosavovic2020designing,wang2024repvit,wang2024yolov10}, recent research has further investigated various ways to augment them from different perspectives for higher efficiency, including training data and model architecture, \etc~\cite{lin2024moe,chu2023mobilevlm}. For example, Bunny~\cite{he2024efficient} constructs informative datasets by curated data selection from broader sources. ALLaVA~\cite{chen2024allava} leverages strong proprietary models to generate both fine-grained image annotations and complex reasoning visual question-answering pairs. LLaVA-Phi~\cite{zhu2024llavaphi} and TinyGPT-V~\cite{yuan2023tinygpt} utilize the Phi-2~\cite{javaheripi2023phi} LLM backbone to advance the realm of compact MLLMs. MOE-LLaVA~\cite{lin2024moe} explore the mixture of experts to enhance the model performance with minimal inference overhead. Moreover, MLLMs exhibit a broad range of application prospects, including autonomous driving~\cite{cui2024survey,ding2023hilm,wang2023drivemlm,xu2024drivegpt4}, intelligent robotics~\cite{wang2024large,wang2023drivemlm,long2024robollm,driess2023palm}, retrieval system~\cite{chen2024mllm,yue2024object,wang2023hierarchical,ding2023exploring}, and smart healthcare~\cite{li2024llava,mesko2023impact,xiao2024comprehensive,alsaad2024multimodal}, \etc.

\textbf{Visual Token Compression.}
Despite showing strong multimodal understanding and generation capabilities, the introduction of visual information also brings long visual token sequences. For example, mini-Gemini-HD~\cite{li2024mini} and Fuyu~\cite{fuyu-8b} transform pixel-level images of 672$\times$672 and 1024$\times$1024 to 2880 and 1296 tokens, respectively. The large number of visual tokens leads to the notable computation and memory overhead, resulting in a bottleneck for inference speed of MLLMs. To address this, like research on token compression for Vision Transformers~\cite{dosovitskiy2020image,bolya2022token,wang2023cait}, recent works have investigated various compression ways from both the connector architecture design and visual token pruning perspectives for MLLMs. For example, VoCo-LLaMA~\cite{ye2024voco} utilizes the LLM's understanding paradigm of visual tokens to compress them into the special VoCo tokens. DeCo~\cite{yao2024deco} introduces the adaptive pooling operation to squeeze visual tokens by downsampling. LLaVA-PruMerge~\cite{shang2024llava} adaptively reserves the most crucial visual tokens and merges the unimportant ones with them based on the key similarity. DeepStack~\cite{meng2024deepstack} proposes to stack the visual tokens into several groups and feed them to the transformer layers of LLMs from bottom to top. However, these methods generally require substantial resources to train new architecture, or suffer from the significant perform degradation due to suboptimal token pruning. Effective and efficient visual compression methods are essential to fully harness the potential of MLLMs and facilitate their applications.

LLaVA-PruMerge~\cite{shang2024llava} also reduce the visual token before the process of LLM, leveraging the [CLS] token. They design a complex merging framework which requires fine-tuning with training dataset. In contrast, we advocate a training-free eviction-based compression of visual tokens purely based on its relationship with [CLS] token. We also propose to ensemble the importance estimation across layers to enhance the preservation of comprehensive visual information for LLM's prompting and generation. Experimental analysis shows that our method can outperform the training-free variant of LLaVA-PruMerge.

\section{Methodology}
In this section, we present our VTC-CLS method for visual token compression of MLLMs. We first introduce the token importance indicator in visual encoder and the one on LLM of MLLMs in \Cref{sec:preliminary}. We then reveal the high consistency between important visual tokens in the visual encoder, \ie, those with high attention from [CLS] token, and the important ones derived by attention in LLM, \ie, those with high attention from output textual tokens, in \Cref{sec:consistency}. Based on this, in \Cref{sec:vtc-cls}, we propose to utilize the attention of [CLS] token on visual tokens as the indicator for importance to perform visual token pruning for MLLMs.

\begin{figure*}[t]
    \centering
    \includegraphics[width=0.85\linewidth]{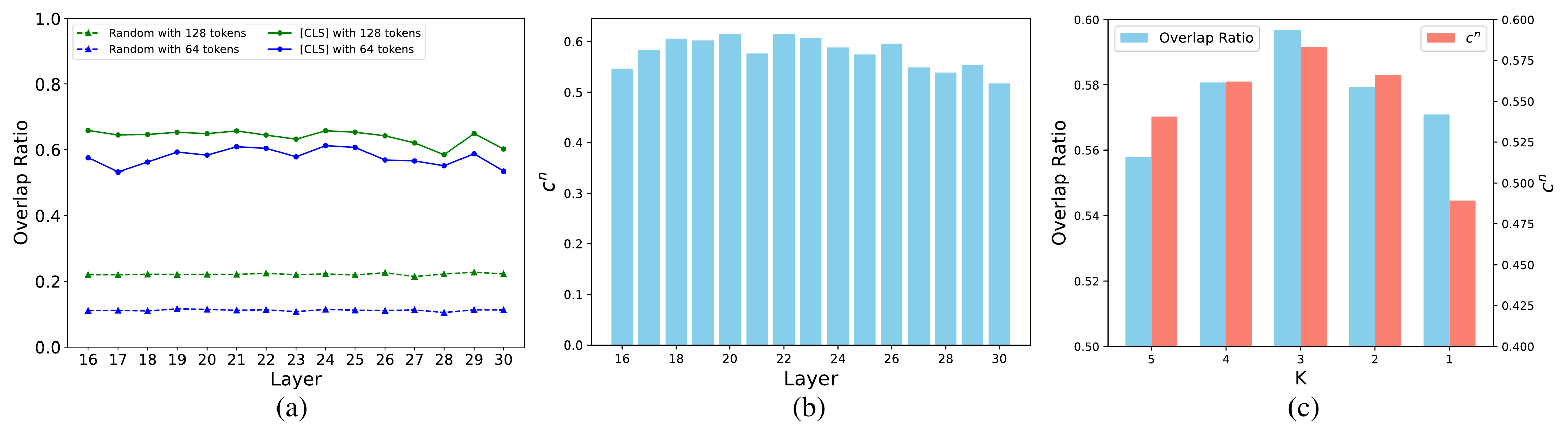}
    \caption{(a) Overlap proportion of important visual tokens in the visual encoder and LLM's layers. We leverage the attention scores of [CLS] token and random values as the token importance score in visual encoder for inspection, denoted as ``[CLS]'' and ``Random'', respectively. Under the kept visual token number of 128 and 64, ``[CLS]'' can stably show the high overlap ratio, indicating the high consistency with the importance score in LLM. (b) The spearman's rank correlation coefficient of importance scores in the visual encoder and LLM's layers. (c) Average overlap ratio of important tokens and spearman's rank correlation coefficient of importance scores in the visual encoder and LLM's layers under different $K$. It shows that ensembling the importance across layers in visual encoder can strength its consistency with that of LLM. The best K is 3.}
    \label{fig:cls-llm-overlap}
    \vspace{-10pt}
\end{figure*}

\subsection{Preliminary}
\label{sec:preliminary}
MLLMs typically comprise of three modules, \ie, the visual encoder, the connector, and the LLM backbone. 

\textbf{Token importance in visual encoder.} 
Benefiting from the pretrained on a large number of image-text pairs, CLIP-ViT~\cite{radford2021learning} becomes the common choice for visual encoder. It embeds the input image as several patches and concatenates an additional [CLS] token for aligning with the textual representation. Multiple transformer layers are stacked for processing the patches, which interact with each other through the self-attention mechanism. Suppose that the head number of the self-attention module is $H_v$, for $i$-th head, we denote the corresponding key and value features as $\boldsymbol{K}_v^i\in \mathbb{R}^{(N_v+1)\times D_v}$, and $\boldsymbol{V}_v^i \in \mathbb{R}^{(N_v+1)\times D_v}$, respectively. $N_v$ is the number of visual patches and $D_v$ is the head dimension. In the $i$-th head, the feature $\boldsymbol{y}^{i}_v$ of [CLS] token is updated by
\begin{equation}
    \small
    \boldsymbol{y}^{i}_v = \boldsymbol{a}^{i}_v\boldsymbol{V}_v^i = \text{softmax}(\frac{\boldsymbol{q}^{i}_v {\boldsymbol{K}_v^i}^T}{\sqrt{D_v}})\boldsymbol{V}_v^i,
\end{equation}where $\boldsymbol{q}_v^{i}$ denotes the query feature of the [CLS] token and $\boldsymbol{a}_v^{i}$ means the attention score between the [CLS] token and the visual tokens. Considering that [CLS] token aggregates the overall image information, a higher correlation with the [CLS] token indicates that visual token contains more key information~\cite{liang2022not}. Therefore, $\boldsymbol{a}_v^{i}$ can be employed as an effective metric for evaluating visual token importance in the visual encoder. For multiple heads, we average the attention scores by
\begin{equation}
    \small
    \label{eq:init-visual-importance}
    \boldsymbol{s}_v = \frac{\sum_i \boldsymbol{a}_v^i}{H_v},
\end{equation} where $\boldsymbol{s}_v$ is the final importance 
score of visual tokens.

\textbf{Token importance in the LLM of MLLMs.}
After transforming the image into visual token sequences, the connector projects them into the feature space of the textual embeddings. The visual tokens, along with textual tokens including system and question prompts, can thus be concatenated and fed into the LLM for processing. The LLM typically adopts the transformer architecture with the causal self-attention mechanism. We suppose the number of heads in the self-attention module is $H_l$. For the $j$-th head, we denote the key, and value features of the input tokens as $\boldsymbol{K}_l^j \in \mathbb{R}^{(N_l+1)\times D_l}$, and $\boldsymbol{V}_l^j \in \mathbb{R}^{(N_l+1)\times D_l}$, respectively. $N_l$ is the input token number and $D_l$ is the head dimension. The updated feature $\boldsymbol{y}_l^j$ of the output token for generation in the $j$-th head is derived by
\begin{equation}
    \small
    \boldsymbol{y}_l^j=\boldsymbol{a}_l^j \boldsymbol{V}_l^j = \text{softmax}(\frac{\boldsymbol{q}_l^j {\boldsymbol{K}_l^j}^T}{\sqrt{D_l}})\boldsymbol{V}_l^j,
\end{equation}where $\boldsymbol{q}_l^{j}$ denotes the query feature of the output token and $\boldsymbol{a}_l^{j}$ means the attention score between it and others. To achieve accurate multimodal understanding, the output token needs to attend to visual tokens and text tokens for capturing the information from the visual modality and language modality, respectively. The attention score that output tokens assign to visual tokens can thus reflect the their significance in LLM. During inference, LLM generates the response progressively in the autoregressive manner. For the $k$-th output token, we denote its attention score to the visual tokens in the $j$-th head as $\boldsymbol{a}_l^{j,k}$. Suppose that the number of output tokens is $O$, the attention scores can thus be averaged across all heads and output tokens as the importance indicator $\boldsymbol{s}_l$ of visual tokens in the LLM by
\begin{equation}
    \small
    \label{eq:llm-importance}
    \boldsymbol{s}_l = \frac{\sum_j\sum_k \boldsymbol{a}_l^{j,k}}{H_l\cdot O}.
\end{equation}

\begin{figure*}[t]
    \centering
    \includegraphics[width=0.75\linewidth]{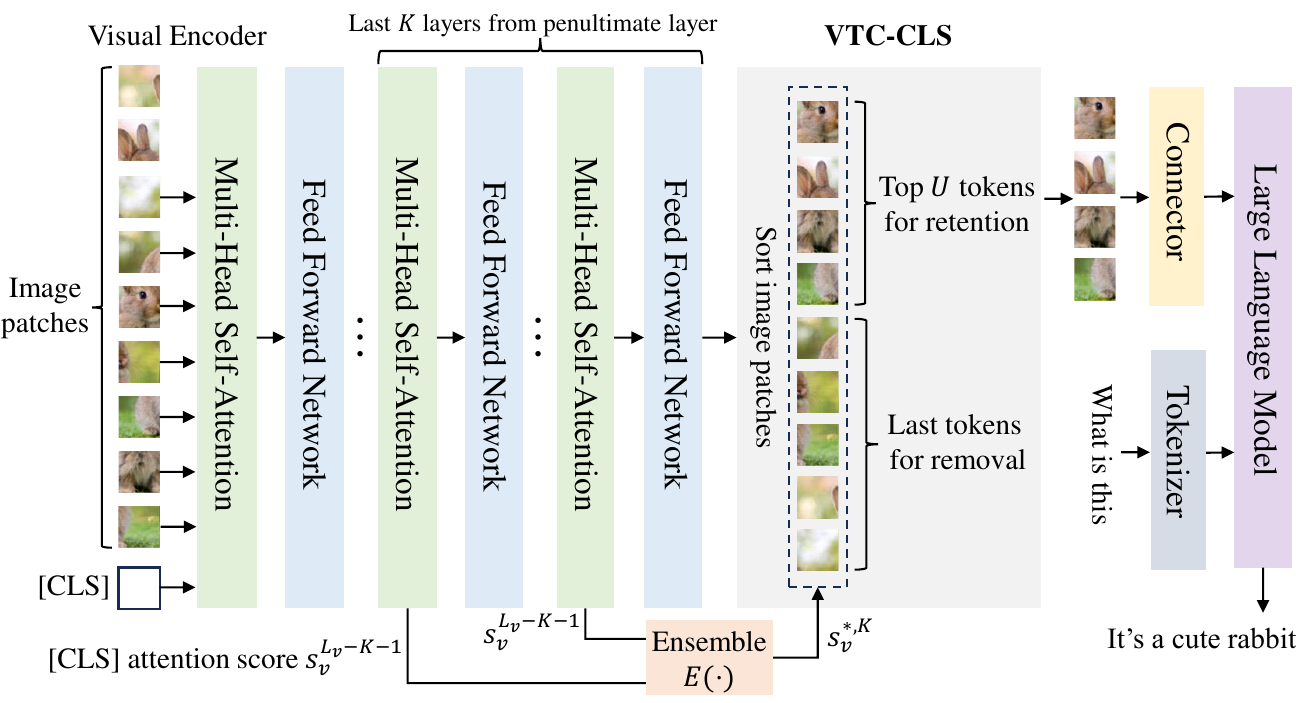}
    \caption{The pipeline of our method. Motivated by the high consistency between the important visual tokens in the visual encoder and those in LLM of MLLMs, we leverage the attention scores of the [CLS] token in the visual encoder as the importance indicator. We ensemble the importance scores across different layers of visual encoder for joint selection and reserve the critical ones needed by LLM.}
    \label{fig:pipeline}
    \vspace{-10pt}
\end{figure*}

\subsection{Consistency of Visual Token Importance}
\label{sec:consistency}
We note that in practice, the output tokens for the response are unknown in advance, making it difficult to accurately determine the important visual tokens, \ie, obtaining $\boldsymbol{s}_l$, for LLMs. Therefore, we approach from the visual encoder perspective, and investigate the relationship between visual token importance in visual encoder and the LLM. Suppose that the visual encoder consists of $L_v$ transformer layers, for the $m$-th layer, we denote the importance score of visual tokens by $\boldsymbol{s}_v^m$ according to \Cref{eq:init-visual-importance}. Considering that the LLM usually leverages the features of penultimate layer in visual encoder~\cite{liu2024visual,liu2024improved}, we thus adopt the corresponding $\boldsymbol{s}_v^{L_v-1}$ as the importance metric of visual tokens in visual encoder. For the LLM with $L_{l}$ transformer layers, similarly, we denote the importance score of visual tokens by $\boldsymbol{s}_l^n$ for the $n$-th LLM layer based on \Cref{eq:llm-importance}. 

We then obtain random image-text pairs to collect $\boldsymbol{s}_v^{L_v-1}$ and $\boldsymbol{s}_l^n$ for inspection. Given a target visual token number of $U$ after compression, we can derive the important visual token set $\Omega_v$ in visual encoder according to $\boldsymbol{s}_v^{L_v-1}$. For LLM, we can also derive the important visual token set $\Omega_l^n$ for the $n$-th layer based on $\boldsymbol{s}_l^{n}$. The process can be formulated by
\begin{equation}
    \small
    \Omega_v = \text{TopU}({\boldsymbol{s}_v^{L_v-1}});\; \Omega_l^n = \text{TopU}(\boldsymbol{s}_l^{n}).
\end{equation} To explore the association between $\boldsymbol{s}_v^{L_v-1}$ and $\boldsymbol{s}_l^n$, we calculate the overlap proportion $p^n$ of important visual tokens in the visual encoder and LLM by
\begin{equation}
    \small
    p^n = \frac{|\Omega_v \cap \Omega_l^{n}|}{U}.
\end{equation} As shown in \Cref{fig:cls-llm-overlap}.(a), we observe that the overlap ratio of important tokens in the visual encoder and LLM's layers is consistently high under various $U$, indicating the obvious consistency. The similar important tokens between visual encoder and LLM means the similar importance scores, suggesting their potential positive correlation. To verify that such high consistency is attributed to the utilization of [CLS] token, we also inspect the overlap ratio of random importance scores and the importance score in LLM's layers. As shown in \Cref{fig:cls-llm-overlap}.(a), ``Random'' exhibits much lower consistency, verifying the effectiveness of [CLS] token. These results inspire that we can approximate the visual token importance in LLM using the scores based on [CLS] token from the visual encoder perspective.

We leverage spearman's rank correlation coefficient~\cite{sedgwick2014spearman} to quantitatively assess the positive correlation of importance scores between the visual encoder and the LLM, \ie, the degree of approximation. Specifically, the rank of each visual token based on its important score directly influences its selection during the compression process. If the rank of each visual token in visual encoder matches its rank in the LLM, then the tokens selected based on the importance scores from visual encoder will match the optimal selections based on the LLM's importance score. Therefore, for each visual token $u$, we first derive its ranks $\boldsymbol{r}_v^u$ and $\boldsymbol{r}_l^{u,n}$ in the visual encoder and the $n$-th layer of LLM according to $\boldsymbol{s}_v^{L_v-1}$ and $\boldsymbol{s}_l^n$, respectively, by
\begin{equation}
    \small
    \boldsymbol{r}_v^u=\text{rank}_{\boldsymbol{s}_v^{L_v-1}}(u);\quad \boldsymbol{r}_l^{u,n}=\text{rank}_{\boldsymbol{s}_l^n}(u).
\end{equation} After obtaining the rank sequence $\{\boldsymbol{r}_v^1, \boldsymbol{r}_v^2,..., \boldsymbol{r}_v^{N_v}\}$ of the visual encoder and $\{\boldsymbol{r}_l^{1, n}, \boldsymbol{r}_v^{2, n},..., \boldsymbol{r}_v^{N_v, n}\}$ of the $n$-th LLM layer, we can thus calculate their pearson correlation coefficient $c^n$ for evaluating the spearman's rank correlation coefficient between importance scores $\boldsymbol{s}_v^{L_v-1}$ and $\boldsymbol{s}_l^n$~\cite{cohen2009pearson}. The process can be formulated by
\begin{equation}
    \small
    c^n=\frac{\sum_u(\boldsymbol{r}_v^u - \overline{\boldsymbol{r}}_v)(\boldsymbol{r}_l^{u,n}-\overline{\boldsymbol{r}}_l^{n})}{\sqrt{\sum_u(\boldsymbol{r}_v^u-\overline{\boldsymbol{r}}_v)}\sqrt{\sum_u(\boldsymbol{r}_l^{u,n}-\overline{\boldsymbol{r}}_l^{n}})}. 
\end{equation} As $c^n$ increases, it signifies a stronger positive correlation between $\boldsymbol{s}_v^{L_v-1}$ and $\boldsymbol{s}_l^n$, indicating more accurate approximation of the importance scores in visual encoder for those in LLM. As shown in \Cref{fig:cls-llm-overlap}.(b), the spearman's rank correlation coefficient between visual encoder and LLM's layers shows consistently high trends. This means that the importance score of visual tokens in the visual encoder is comparable to their importance in the LLM, indicating the strong positive correlation. It suggests the reasonableness of estimating visual token importance and reserving the crucial ones for LLM from the visual encoder perspective.

\subsection{Visual Token Compression by [CLS] Token}
\label{sec:vtc-cls}
The above observation motivates us to perform visual token compression based on the $\boldsymbol{s}_v^{L_v-1}$, \ie, the attention scores of [CLS] token, in the visual encoder. Specifically, given an image-text pair, for visual tokens, we obtain the average attention scores of the [CLS] token across heads in the penultimate layer as their importance indicators, \ie, $\boldsymbol{s}_v^{L_v-1}$. Under any target visual token number $U$ after compression, the top $U$ visual tokens based on $\boldsymbol{s}_v^{L_v-1}$ are reserved and the left tokens are removed. The retained visual tokens are then concatenated with textual tokens and fed into LLM for inference. In this way, the key visual information can be effectively attended to in LLM, with notable less computation and memory overhead, as shown in \Cref{fig:pipeline}.

\textbf{Ensembling importance across layers.}
Due to the variation of scale of different objects in the image, in different layers of the visual encoder like CLIP-ViT, the [CLS] token may attend to different salient object regions to more comprehensively capture the content of the entire image~\cite{walmer2023teaching}. Therefore, solely leveraging the [CLS] token of penultimate layer in visual encoder may provide a biased importance estimation of visual tokens. Recognizing this, we further explore the joint importance assessment by ensembling the prior knowledge of token importance in different layers.

\begin{table*}[t!]
    \centering
    \caption{Comparisons with the state-of-the-arts across various visual-language benchmarks.}
    \label{tab:result}
    \setlength{\tabcolsep}{5pt}
    \begin{tabular}{l|ccccccccc}
    \toprule
    Model & SciQA & GQA & TextVQA & POPE & MMB & MMB-CN & MMVet & SEED & Avg \\
    \midrule
    BLIP2-14B~\cite{li2023blip} & 61.0 & 41.0 & 42.5 & 85.3 & - & - & 22.4 & 49.7 & - \\
    InstructBLIP-14B~\cite{dai2023instructblip} & 63.1 & 49.5 & 50.7 & 78.9 & - & - & 25.6 & - & - \\
    IDEFICS-80B~\cite{idefics} & - & 45.2 & 30.9 & - & 54.5 & 38.1 & - & 53.2 & - \\
    Qwen-VL-Chat~\cite{bai2023qwen} & 68.2 & 57.5 & 61.5 & - & 60.6 & 56.7 & - & 65.4 & - \\
    \midrule
    \rowcolor[gray]{0.92}
    \multicolumn{10}{c}{Results with 576 visual tokens} \\
    \specialrule{0em}{0.8pt}{0.8pt}
    LLaVA-1.5-7B~\cite{liu2024improved} & 70.2 & 62.0 & 58.3 & 85.9 & 64.7 & 58.1 & 31.1 & 66.2 & 62.1 \\
    \midrule
    \rowcolor[gray]{0.92}
    \multicolumn{10}{c}{Results with 256 visual tokens} \\
    \specialrule{0em}{0.8pt}{0.8pt}
    FastV~\cite{chen2024image} & 69.3 & 59.1 & 58.0 & 80.2 & 64.1 & 57.8 & 30.9 & 64.5 & 60.5 \\
    PruMerge+~\cite{shang2024llava} & 70.0 & 59.5 & 55.1 & 84.1 & 63.2 & 57.2 & 31.8 & 64.0 & 60.6 \\
    \textbf{VTC-CLS} & 69.6 & 60.3 & 57.7 & 86.2 & 63.6 & 57.1 & 34.4 & 64.8 & \textbf{61.7} \\
    \midrule
    \rowcolor[gray]{0.92}
    \multicolumn{10}{c}{Results with 128 visual tokens} \\
    \specialrule{0em}{0.8pt}{0.8pt}
    FastV~\cite{chen2024image} & 69.3 & 54.1 & 56.4 & 67.6 & 63.1 & 55.9 & 26.7 & 59.6 & 56.6 \\
    PruMerge+~\cite{shang2024llava} & 70.1 & 57.8 & 54.3 & 81.5 & 61.3 & 54.7 & 28.7 & 60.8 & 58.6 \\
    \textbf{VTC-CLS} & 69.7 & 58.2 & 56.8 & 84.0 & 62.5 & 56.8 & 33.4 & 62.2 & \textbf{60.5} \\
    \midrule
    \rowcolor[gray]{0.92}
    \multicolumn{10}{c}{Results with 64 visual tokens} \\
    \specialrule{0em}{0.8pt}{0.8pt}
    FastV~\cite{chen2024image} & 68.6 & 46.0 & 51.4 & 35.6 & 50.4 & 42.3 & 21.2 & 47.1 & 45.3 \\
    PruMerge+~\cite{shang2024llava} & 70.1 & 55.9 & 53.0 & 77.4 & 59.3 & 51.0 & 25.9 & 58.3 & 56.3 \\
    \textbf{VTC-CLS} & 70.0 & 55.7 & 55.5 & 78.8 & 61.2 & 55.8 & 31.7 & 58.3 & \textbf{58.4} \\
    \bottomrule
    \end{tabular}
    \vspace{-10pt}
\end{table*}

Specifically, considering that the information aggregated by the [CLS] token becomes more thorough in the later layers~\cite{walmer2023teaching}, we select $K$ layers starting from the penultimate layer and moving backward, and obtain the corresponding importance scores of $\{\boldsymbol{s}_v^{L_v-K}, \boldsymbol{s}_v^{L_v-K+1},..., \boldsymbol{s}_v^{L_v-1}\}$ based on the attention scores of [CLS] token. Then, an ensemble function $E(\cdot)$ is employed to aggregate them, \ie,
\begin{equation}
    \small
    \label{eq:visual-importance}
    \overline{\boldsymbol{s}}_v^{K}=E(\{\boldsymbol{s}_v^{L_v-K}, \boldsymbol{s}_v^{L_v-K+1},..., \boldsymbol{s}_v^{L_v-1}\}),
\end{equation} where $\overline{\boldsymbol{s}}_v^{K}$ denotes the final importance scores of visual tokens. We employ the average operation for the ensemble function by default and inspect spearman's rank correlation coefficient between $\overline{\boldsymbol{s}}_v^{K}$ and the importance metric $\boldsymbol{s}_l^{n}$ in LLM under different $K$. As shown in \Cref{fig:cls-llm-overlap}.(c), compared with only leveraging $\boldsymbol{s}_v^{L_v-1}$, \ie, $K=1$, adopting $K$ from 2 to 5 can consistently obtain the higher overlap ratio and spearman's rank correlation coefficient. This shows that ensembling the importance scores from multiple layers of visual encoder can enhance the positive correlation with those in LLM. We observe that the strongest correlation can be achieved under $K=3$ and thus adopt this by default.

\section{Experiments}

\subsection{Implementation Details}
Following~\cite{shang2024llava,chen2024image}, we verify the proposed VTC-CLS on LLaVA-1.5~\cite{liu2024improved}. It leverages the ViT-L~\cite{dosovitskiy2020image} with the published CLIP weights as the visual encoder and the Vicuna v1.5~\cite{chiang2023vicuna,zheng2023judging} as the LLM backbone for multimodal generation. We evaluate the performance with our visual token compression method on eight widely used benchmarks, including ScienceQA~\cite{lu2022learn}, TextVQA~\cite{singh2019towards}, MMBench~\cite{liu2023mmbench}, MMBench-CN~\cite{liu2023mmbench}, POPE~\cite{li2023evaluating}, MMVet~\cite{yu2023mm}, GQA~\cite{hudson2019gqa}, and SEED-Bench~\cite{li2024seed}, in a training-free manner. Specifically, the image subset of ScienceQA is used to assess the zero-shot generalization on scientific question answering. TextVQA consists of text-rich visual question answering and evaluate the MLLM's fined-grained visual perception capability. MMBench and MMBench-CN leverage all-round shuffling on multiple choice answers to assess the MLLM's answer robustness. POPE adopts three-sampled subsets of COCO~\cite{lin2014microsoft}, including random, common, and adversarial, to evaluate the MLLM's degree of hallucination, and the F1 score is reported on all three splits. MMVet measures the MLLM's capabilities in engaging in visual conversations on a diverse range of tasks. GQA tests the MLLM for understanding the visual scenes and reasoning about various aspects of images. SEED-Bench introduces the comprehensive evaluation for MLLMs across 12 dimensions across both spatial and temporal understanding. We empirically set $K$ to 3 for VTC-CLS, by default. All methods are adopted in a training-free way for fair comparisons.

\subsection{Main Results}
As shown in \Cref{tab:result}, our VTC-CLS achieves the state-of-the-art performance under various compression levels. Specifically, with 256 visual tokens fed into LLM, VTC-CLS obtains the notable overall improvements of 1.2\% and 1.1\% compared with FastV and PruMerge+, respectively. When retaining only 64 visual tokens, our VTC-CLS also significantly outperforms PruMerge+ by 2.1\% on average. The improvements are evident across various capabilities of MLLMs. For example, with 128 visual tokens retained, VTC-CLS surpasses PruMerge+ by 2.5\% on POPE, showing much less hallucination. It also improves 4.1\% over FastV on GQA, indicating the better visual scene understanding capability. We also note that our method well maintain the competitive performance compared with the original model. For example, it only leads to 0.4\% average performance degradation when more than half of the tokens are compressed. These results show the effectiveness of our method, which can well preserve the MLLM's abilities. Furthermore, despite having notably fewer visual tokens, our method also performs better than other MLLMs, like Qwen-VL-Chat and InstructBLIP, \etc.

\begin{table}[t!]
    \centering
    \setlength{\tabcolsep}{3pt}
    \caption{Inference efficiency analyses based on LLaVA-1.5-7B.We measure the evaluation time on the test dataset.}
    \label{tab:efficiency}
    \begin{tabular}{c|c|cc|cc}
    \toprule
    \multirow{2}{*}{Method} & Token  &\multicolumn{2}{c|}{GQA} & \multicolumn{2}{c}{POPE} \\
    \cmidrule{3-6}
    & Number & Time (s) & Acc. (\%) & Time (s)& F1 (\%) \\
    \midrule
    \specialrule{0em}{0.8pt}{0.8pt}
    Base & 576 & 3018 & 62.0 & 1921 & 85.9 \\
    \midrule
    \specialrule{0em}{0.8pt}{0.8pt}
    \multirow{2}{*}{Ours} & 128 & 2049 & 58.2  & 1349 & 84.0  \\
    \specialrule{0em}{0.8pt}{0.8pt}
     & 64 & 1750 & 55.7 & 1216 & 78.8 \\
    \bottomrule
    \end{tabular}
\end{table}

\subsection{Model analyses}
We present comprehensive analyses for our method, providing deeper insights. Experiments are conducted on LLaVA-1.5-7B across various benchmarks including GQA, POPE, MMVet, and SEED-Bench, by default.

\textbf{Inference efficiency analyses.} To investigate the benefit of inference efficiency brought by our method, we measure the total of evaluation time during MLLM inference with and without the visual token compression. We conduct analyses on a single NVIDIA 3090 GPU. As shown in \Cref{tab:efficiency}, we observe that our method can accelerate the inference of MLLMs notably across different datasets under distinct compression levels. With 128 visual tokens retained, VTC-CLS reduces the total time by 32.1\% and 29.8\% on GQA and POPE, respectively. Besides, it consistently maintains competitive performance compared with the original model. Furthermore, when reserving only 64 visual tokens, VTC-CLS provides notable 1.7$\times$ and 1.6$\times$ speedups in two scenarios, respectively. These results demonstrate the high practicality of our method, showing its favorable efficiency enhancement.

\begin{table}[t!]
    \centering
    \caption{Effect of importance score based on [CLS] token.}
    \label{tab:cls}
    \begin{tabular}{l|ccccc}
    \toprule
    Method &  GQA & POPE & MMVet & SEED & Avg \\
    \midrule
    Random & 57.3 & 82.2 & 29.5 & 60.9 & 57.5 \\
    ToMe & 58.0 & 85.6 & 29.7 & 60.7 & 58.5 \\
    KMeans & 58.1 & 83.9 & 30.1 & 61.8 & 58.5 \\
    \textbf{VTC-CLS} & 58.2 & 84.0 & 33.4 & 62.2 & \textbf{59.5}  \\
    \bottomrule
    \end{tabular}
    \vspace{-10pt}
\end{table}

\textbf{Effect of importance score based on [CLS] token.} We verify the effectiveness of assessing the visual token importance and retaining important ones according to the [CLS] token in the visual encoder. We introduce three baselines:  \begin{enumerate*}[label = (\arabic*)] 
    \item assigning random importance scores to the visual tokens for compression, denoted as ``Random''.
    \item dividing the visual tokens into two sets and merging similar ones by bipartite soft matching following ToMe~\cite{bolya2022token}, denoted as ``ToMe''.
    \item clustering the visual tokens by KMeans~\cite{lloyd1982least,macqueen1967some} with the cluster centers taken as the compressed tokens, denoted as ``KMeans''.
\end{enumerate*}    
We conduct experiments based on LLaVA-1.5-7B with 128 visual tokens retained. As shown in \Cref{tab:cls}, ``Random'' exhibits notably inferior performance compared with VTC-CLS, verifying the effectiveness of the utilization of [CLS] token. Furthermore, ``ToMe'' and ``KMeans'' merge similar visual tokens for compression. While they can achieve good results on certain benchmarks, such as POPE, where token merging helps reduce the hallucination, they still exhibit inferior overall performance compared with ours. This can be attributed to the fact that ``ToMe'' and ``KMeans'' overlook the varying importance of visual tokens for LLM and may disturb the features of important ones due to merging features with discrepancies. In contrast, our method leverages the abundant prior knowledge of [CLS] token in visual encoder, which can precisely select important visual tokens needed by LLM and bring favorable performance.

\begin{table}[t!]
    \centering
    \caption{The impact of different $K$ in VTC-CLS.}
    \label{tab:K}
    \begin{tabular}{l|ccccc}
    \toprule
    $K$ & GQA & POPE & MMVet & SEED & Avg  \\
    \midrule
    1 & 57.9 & 82.6 & 31.6 & 61.8 & 58.5 \\
    2 & 58.0 & 83.6 & 32.2 & 62.1 & 59.0 \\
    3 & 58.2 & 84.0 & 33.4 & 62.2 & 59.5 \\
    4 & 58.0 & 83.9 & 33.0 & 62.2 & 59.3 \\
    5 & 57.9 & 83.6 & 32.5 & 62.0 & 59.0 \\
    \bottomrule
    \end{tabular}
\end{table}

\begin{table}[t!]
    \centering
    \caption{Different ensemble function $E(\cdot)$ in VTC-CLS.}
    \label{tab:ensemble}
    \begin{tabular}{l|ccccc}
    \toprule
    $E(\cdot)$ & GQA & POPE & MMVet & SEED & Avg \\
    \midrule
    none & 57.9 & 82.6 & 31.6 & 61.8 & 58.5 \\
    avg & 58.2 & 84.0 & 33.4 & 62.2 & 59.5 \\
    max & 58.2 & 83.8 & 31.7 & 61.8 & 58.9 \\
    min & 58.3 & 83.8 & 33.0 & 62.0 & 59.3 \\
    \bottomrule
    \end{tabular}
    \vspace{-10pt}
\end{table}

\begin{figure*}[t]
    \centering
    \includegraphics[width=0.8\linewidth]{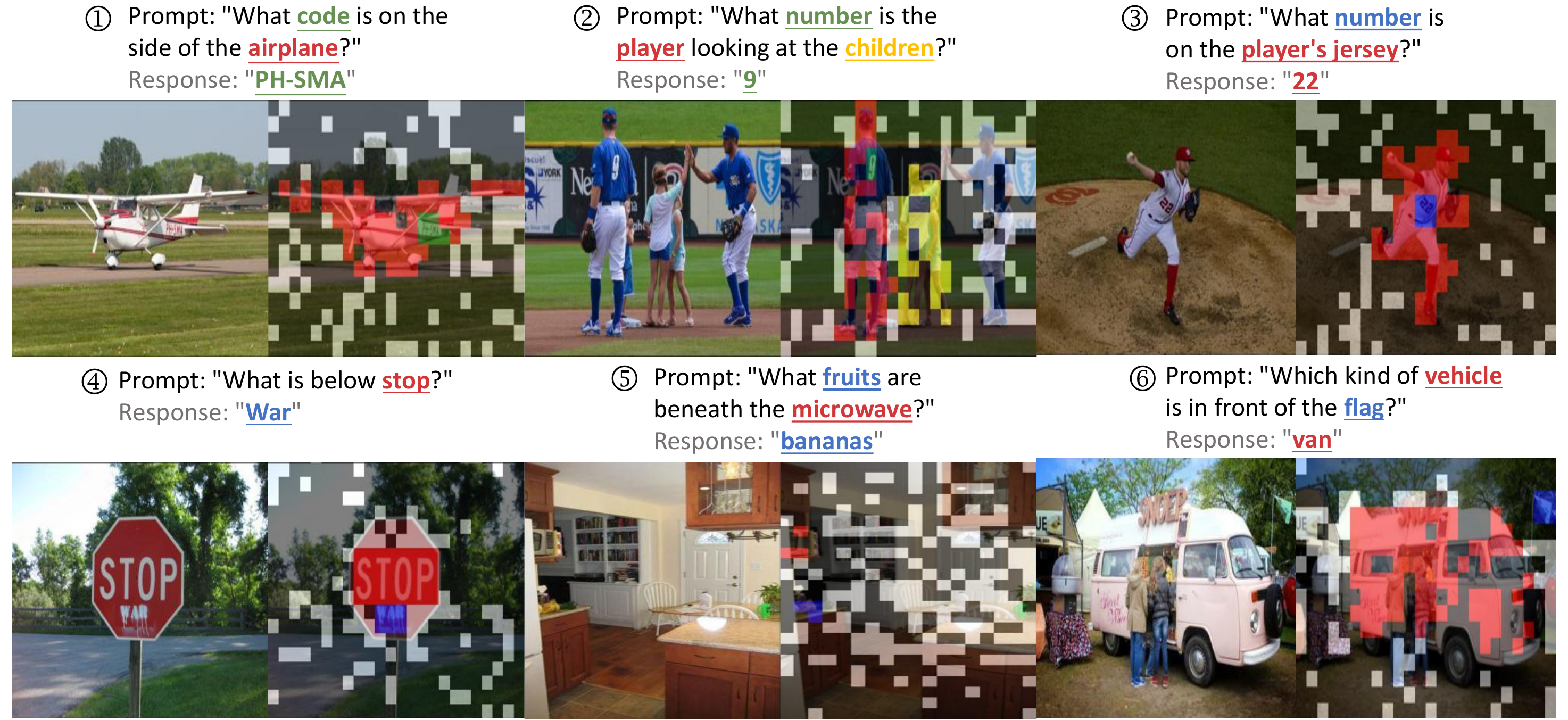}
    \caption{Visualization of retained visual patches. The areas masked in black represent the discarded visual tokens. Besides, we further show the correspondence between the salient visual regions and texts by different colors. It can be observed that our method effectively removes redundant visual signals while preserving the significant ones, enabling various textual tokens to perceive the corresponding visual modality in LLM. Best viewed in color.}
    \label{fig:vis-token}
    \vspace{-15pt}
\end{figure*}

\textbf{Impact of different $K$.} We investigate the performance variation under different values of $K$ in VTC-CLS. We conduct experiments based on LLaVA-1.5-7B with 128 visual tokens retained for inspection. As shown in \Cref{tab:K}, as $K$ gradually increases, the performance exhibits a trend of initially increasing and then subsequently decreasing. The best overall performance can be achieved under the $K$ of 3. We also note that the trend in performance is similar with the spearman's rank correlation coefficient trend in \Cref{fig:cls-llm-overlap}.(c). This finding further indicates the effectiveness of pursuing high consistency with the importance scores of visual tokens in LLM. Moreover, adopting $K$ greater than 1 can consistently achieve better performance compared with the scenario of $K=1$. This shows the effectiveness of ensembling the abundant knowledge across different layers, which considers the key visual information more comprehensively.

\textbf{Different ensemble function.} By default, we leverage the average operation to aggregate the different attention scores of [CLS] tokens across layers. We also conduct experiments using different ensemble function $E(\cdot)$, including the maximum operation and minimum operation, which are denoted as ``max'' and ``min'', respectively. We inspect the performance comparisons with 128 visual tokens retained based on LLaVA-1.5-7B. We also introduce the baseline without ensemble, \ie, setting $K$ to 1, which is denoted as ``none''. As shown in \Cref{tab:ensemble}, different ensemble operations can consistently boost the performance compared with ``none'' across various benchmarks. This shows the effectiveness and robustness of ensembling importance from different layers, indicating that its benefit is not limited to the specific ensemble function. Besides, simply adopting average operation can achieve the best overall performance.

\begin{table}[t!]
    \centering
    \caption{Compression in LLM.}
    \label{tab:compression-llm}
    \begin{tabular}{l|ccccc}
    \toprule
    Method & GQA & POPE & MMVet & SEED & Avg \\
    \midrule
    FastV & 54.3 & 67.6 & 26.7 & 59.6 & 52.1 \\
    \textbf{VTC-CLS} & 58.5 & 81.2 & 32.1 & 62.4 & \textbf{58.6} \\
    \bottomrule
    \end{tabular}
    \vspace{-10pt}
\end{table}

\textbf{Compression in LLM.} We explore the feasibility of applying VTC-CLS for compression in the layer of LLM. Specifically, we follow FastV~\cite{chen2024image} to retain the important visual tokens after the second layer of LLM according to the importance score derived from visual encoder, \ie, $\overline{\boldsymbol{s}}_v^{K}$ in \Cref{eq:visual-importance}. We present the comparison results with FastV in \Cref{tab:compression-llm}. It can observed that our method can also achieve superior performance, demonstrating the notable improvements over FastV across different benchmarks. This verifies the effectiveness of the token compression by the abundant prior knowledge in the visual encoder, which can more accurately assess the visual token importance compared with FastV. Besides, this shows the robustness of our method, whose efficacy is not limited in the compression location.

\subsection{Visualization}
To quantitatively show the effectiveness of our method, we visualize the retained visual tokens in various scenarios in \Cref{fig:intro} (bottom) and \Cref{fig:vis-token}. Firstly, our method can reduce the redundancy of visual tokens and remove those corresponding to the similar areas, like the grassland background. Secondly, our method can reserve critical visual tokens relevant to not only the prompt but also the response text, which facilitates LLM to retrieve the corresponding visual information for response generation. For example, in the fourth example in \Cref{fig:vis-token}, the visual tokens capturing ``stop'' in the prompt and ``war'' to be generated are reserved adequately, which greatly facilitate reasoning of semantic relations between them, \ie, ``below'' in the prompt, in LLM. These results verify the efficacy of our method, highlighting its practicality in real-world applications.

\section{Conclusion}
In this paper, we uncover the fact that the visual encoder tells what visual tokens matter for LLM and explore such valuable prior knowledge to compress visual tokens effectively. To this end, we present VTC-CLS, which leverages the attention scores of [CLS] token on visual tokens as the importance indicator for visual token pruning in MLLMs. We further ensemble the importance scores obtained by [CLS] token from different layers to comprehensively capture the key visual information. Extensive experiments demonstrate that our VTC-CLS achieves the state-of-the-art performance across various benchmarks, along with notable inference acceleration, highlighting the superiority.

{
    \small
    \bibliographystyle{ieeenat_fullname}
    \bibliography{main}

\begin{thebibliography}{73}
\providecommand{\natexlab}[1]{#1}
\providecommand{\url}[1]{\texttt{#1}}
\expandafter\ifx\csname urlstyle\endcsname\relax
  \providecommand{\doi}[1]{doi: #1}\else
  \providecommand{\doi}{doi: \begingroup \urlstyle{rm}\Url}\fi

\bibitem[Alayrac et~al.(2022)Alayrac, Donahue, Luc, Miech, Barr, Hasson, Lenc, Mensch, Millican, Reynolds, et~al.]{alayrac2022flamingo}
Jean-Baptiste Alayrac, Jeff Donahue, Pauline Luc, Antoine Miech, Iain Barr, Yana Hasson, Karel Lenc, Arthur Mensch, Katherine Millican, Malcolm Reynolds, et~al.
\newblock Flamingo: a visual language model for few-shot learning.
\newblock \emph{Advances in neural information processing systems}, 35:\penalty0 23716--23736, 2022.

\bibitem[AlSaad et~al.(2024)AlSaad, Abd-Alrazaq, Boughorbel, Ahmed, Renault, Damseh, and Sheikh]{alsaad2024multimodal}
Rawan AlSaad, Alaa Abd-Alrazaq, Sabri Boughorbel, Arfan Ahmed, Max-Antoine Renault, Rafat Damseh, and Javaid Sheikh.
\newblock Multimodal large language models in health care: Applications, challenges, and future outlook.
\newblock \emph{Journal of Medical Internet Research}, 26:\penalty0 e59505, 2024.

\bibitem[Bai et~al.(2023{\natexlab{a}})Bai, Bai, Chu, Cui, Dang, Deng, Fan, Ge, Han, Huang, et~al.]{bai2023qwenllm}
Jinze Bai, Shuai Bai, Yunfei Chu, Zeyu Cui, Kai Dang, Xiaodong Deng, Yang Fan, Wenbin Ge, Yu Han, Fei Huang, et~al.
\newblock Qwen technical report.
\newblock \emph{arXiv preprint arXiv:2309.16609}, 2023{\natexlab{a}}.

\bibitem[Bai et~al.(2023{\natexlab{b}})Bai, Bai, Yang, Wang, Tan, Wang, Lin, Zhou, and Zhou]{bai2023qwen}
Jinze Bai, Shuai Bai, Shusheng Yang, Shijie Wang, Sinan Tan, Peng Wang, Junyang Lin, Chang Zhou, and Jingren Zhou.
\newblock Qwen-vl: A frontier large vision-language model with versatile abilities.
\newblock \emph{arXiv preprint arXiv:2308.12966}, 2023{\natexlab{b}}.

\bibitem[Bavishi et~al.(2023)Bavishi, Elsen, Hawthorne, Nye, Odena, Somani, and Ta\c{s}\i{}rlar]{fuyu-8b}
Rohan Bavishi, Erich Elsen, Curtis Hawthorne, Maxwell Nye, Augustus Odena, Arushi Somani, and Sa\u{g}nak Ta\c{s}\i{}rlar.
\newblock Introducing our multimodal models, 2023.

\bibitem[Bolya et~al.(2022)Bolya, Fu, Dai, Zhang, Feichtenhofer, and Hoffman]{bolya2022token}
Daniel Bolya, Cheng-Yang Fu, Xiaoliang Dai, Peizhao Zhang, Christoph Feichtenhofer, and Judy Hoffman.
\newblock Token merging: Your vit but faster.
\newblock \emph{arXiv preprint arXiv:2210.09461}, 2022.

\bibitem[Chen et~al.(2024{\natexlab{a}})Chen, Chen, Zhang, Chen, Wu, Zhang, Chen, Li, Wan, and Wang]{chen2024allava}
Guiming~Hardy Chen, Shunian Chen, Ruifei Zhang, Junying Chen, Xiangbo Wu, Zhiyi Zhang, Zhihong Chen, Jianquan Li, Xiang Wan, and Benyou Wang.
\newblock Allava: Harnessing gpt4v-synthesized data for a lite vision-language model.
\newblock \emph{arXiv preprint arXiv:2402.11684}, 2024{\natexlab{a}}.

\bibitem[Chen et~al.(2018)Chen, Ding, Lin, Zhao, and Han]{chen2018show}
Hui Chen, Guiguang Ding, Zijia Lin, Sicheng Zhao, and Jungong Han.
\newblock Show, observe and tell: Attribute-driven attention model for image captioning.
\newblock In \emph{IJCAI}, pages 606--612, 2018.

\bibitem[Chen et~al.(2020)Chen, Ding, Liu, Lin, Liu, and Han]{chen2020imram}
Hui Chen, Guiguang Ding, Xudong Liu, Zijia Lin, Ji Liu, and Jungong Han.
\newblock Imram: Iterative matching with recurrent attention memory for cross-modal image-text retrieval.
\newblock In \emph{Proceedings of the IEEE/CVF conference on computer vision and pattern recognition}, pages 12655--12663, 2020.

\bibitem[Chen et~al.(2024{\natexlab{b}})Chen, Zhao, Liu, Bai, Lin, Zhou, and Chang]{chen2024image}
Liang Chen, Haozhe Zhao, Tianyu Liu, Shuai Bai, Junyang Lin, Chang Zhou, and Baobao Chang.
\newblock An image is worth 1/2 tokens after layer 2: Plug-and-play inference acceleration for large vision-language models.
\newblock \emph{arXiv preprint arXiv:2403.06764}, 2024{\natexlab{b}}.

\bibitem[Chen et~al.(2024{\natexlab{c}})Chen, Xu, Qi, and Guo]{chen2024mllm}
Zhanpeng Chen, Chengjin Xu, Yiyan Qi, and Jian Guo.
\newblock Mllm is a strong reranker: Advancing multimodal retrieval-augmented generation via knowledge-enhanced reranking and noise-injected training.
\newblock \emph{arXiv preprint arXiv:2407.21439}, 2024{\natexlab{c}}.

\bibitem[Chiang et~al.(2023)Chiang, Li, Lin, Sheng, Wu, Zhang, Zheng, Zhuang, Zhuang, Gonzalez, et~al.]{chiang2023vicuna}
Wei-Lin Chiang, Zhuohan Li, Zi Lin, Ying Sheng, Zhanghao Wu, Hao Zhang, Lianmin Zheng, Siyuan Zhuang, Yonghao Zhuang, Joseph~E Gonzalez, et~al.
\newblock Vicuna: An open-source chatbot impressing gpt-4 with 90\%* chatgpt quality.
\newblock \emph{See https://vicuna. lmsys. org (accessed 14 April 2023)}, 2\penalty0 (3):\penalty0 6, 2023.

\bibitem[Chu et~al.(2023)Chu, Qiao, Lin, Xu, Yang, Hu, Wei, Zhang, Zhang, Wei, et~al.]{chu2023mobilevlm}
Xiangxiang Chu, Limeng Qiao, Xinyang Lin, Shuang Xu, Yang Yang, Yiming Hu, Fei Wei, Xinyu Zhang, Bo Zhang, Xiaolin Wei, et~al.
\newblock Mobilevlm: A fast, reproducible and strong vision language assistant for mobile devices.
\newblock \emph{arXiv preprint arXiv:2312.16886}, 2023.

\bibitem[Chu et~al.(2024)Chu, Qiao, Zhang, Xu, Wei, Yang, Sun, Hu, Lin, Zhang, et~al.]{chu2024mobilevlm}
Xiangxiang Chu, Limeng Qiao, Xinyu Zhang, Shuang Xu, Fei Wei, Yang Yang, Xiaofei Sun, Yiming Hu, Xinyang Lin, Bo Zhang, et~al.
\newblock Mobilevlm v2: Faster and stronger baseline for vision language model.
\newblock \emph{arXiv preprint arXiv:2402.03766}, 2024.

\bibitem[Cohen et~al.(2009)Cohen, Huang, Chen, Benesty, Benesty, Chen, Huang, and Cohen]{cohen2009pearson}
Israel Cohen, Yiteng Huang, Jingdong Chen, Jacob Benesty, Jacob Benesty, Jingdong Chen, Yiteng Huang, and Israel Cohen.
\newblock Pearson correlation coefficient.
\newblock \emph{Noise reduction in speech processing}, pages 1--4, 2009.

\bibitem[Cui et~al.(2024)Cui, Ma, Cao, Ye, Zhou, Liang, Chen, Lu, Yang, Liao, et~al.]{cui2024survey}
Can Cui, Yunsheng Ma, Xu Cao, Wenqian Ye, Yang Zhou, Kaizhao Liang, Jintai Chen, Juanwu Lu, Zichong Yang, Kuei-Da Liao, et~al.
\newblock A survey on multimodal large language models for autonomous driving.
\newblock In \emph{Proceedings of the IEEE/CVF Winter Conference on Applications of Computer Vision}, pages 958--979, 2024.

\bibitem[Dai et~al.(2023)Dai, Li, Li, Tiong, Zhao, Wang, Li, Fung, and Hoi]{dai2023instructblip}
Wenliang Dai, Junnan Li, Dongxu Li, Anthony Meng~Huat Tiong, Junqi Zhao, Weisheng Wang, Boyang Li, Pascale Fung, and Steven Hoi.
\newblock Instructblip: Towards general-purpose vision-language models with instruction tuning.
\newblock \emph{arXiv preprint arXiv:2305.06500}, 2023.

\bibitem[Ding et~al.(2023{\natexlab{a}})Ding, Han, Xu, Zhang, and Li]{ding2023hilm}
Xinpeng Ding, Jianhua Han, Hang Xu, Wei Zhang, and Xiaomeng Li.
\newblock Hilm-d: Towards high-resolution understanding in multimodal large language models for autonomous driving.
\newblock \emph{arXiv preprint arXiv:2309.05186}, 2023{\natexlab{a}}.

\bibitem[Ding et~al.(2023{\natexlab{b}})Ding, Wang, Chen, Zhang, Liu, Bao, Yan, and Han]{ding2023exploring}
Zixuan Ding, Ao Wang, Hui Chen, Qiang Zhang, Pengzhang Liu, Yongjun Bao, Weipeng Yan, and Jungong Han.
\newblock Exploring structured semantic prior for multi label recognition with incomplete labels.
\newblock In \emph{Proceedings of the IEEE/CVF Conference on Computer Vision and Pattern Recognition}, pages 3398--3407, 2023{\natexlab{b}}.

\bibitem[Dosovitskiy(2020)]{dosovitskiy2020image}
Alexey Dosovitskiy.
\newblock An image is worth 16x16 words: Transformers for image recognition at scale.
\newblock \emph{arXiv preprint arXiv:2010.11929}, 2020.

\bibitem[Driess et~al.(2023)Driess, Xia, Sajjadi, Lynch, Chowdhery, Ichter, Wahid, Tompson, Vuong, Yu, et~al.]{driess2023palm}
Danny Driess, Fei Xia, Mehdi~SM Sajjadi, Corey Lynch, Aakanksha Chowdhery, Brian Ichter, Ayzaan Wahid, Jonathan Tompson, Quan Vuong, Tianhe Yu, et~al.
\newblock Palm-e: An embodied multimodal language model.
\newblock \emph{arXiv preprint arXiv:2303.03378}, 2023.

\bibitem[Dubey et~al.(2024)Dubey, Jauhri, Pandey, Kadian, Al-Dahle, Letman, Mathur, Schelten, Yang, Fan, et~al.]{dubey2024llama}
Abhimanyu Dubey, Abhinav Jauhri, Abhinav Pandey, Abhishek Kadian, Ahmad Al-Dahle, Aiesha Letman, Akhil Mathur, Alan Schelten, Amy Yang, Angela Fan, et~al.
\newblock The llama 3 herd of models.
\newblock \emph{arXiv preprint arXiv:2407.21783}, 2024.

\bibitem[GLM et~al.(2024)GLM, Zeng, Xu, Wang, Zhang, Yin, Rojas, Feng, Zhao, Lai, et~al.]{glm2024chatglm}
Team GLM, Aohan Zeng, Bin Xu, Bowen Wang, Chenhui Zhang, Da Yin, Diego Rojas, Guanyu Feng, Hanlin Zhao, Hanyu Lai, et~al.
\newblock Chatglm: A family of large language models from glm-130b to glm-4 all tools.
\newblock \emph{arXiv preprint arXiv:2406.12793}, 2024.

\bibitem[Gu et~al.(2024)Gu, Zhu, Zhu, Chen, Tang, and Wang]{gu2024anomalygpt}
Zhaopeng Gu, Bingke Zhu, Guibo Zhu, Yingying Chen, Ming Tang, and Jinqiao Wang.
\newblock Anomalygpt: Detecting industrial anomalies using large vision-language models.
\newblock In \emph{Proceedings of the AAAI Conference on Artificial Intelligence}, pages 1932--1940, 2024.

\bibitem[He et~al.(2024)He, Liu, Wu, Yuan, Wang, Huang, and Zhao]{he2024efficient}
Muyang He, Yexin Liu, Boya Wu, Jianhao Yuan, Yueze Wang, Tiejun Huang, and Bo Zhao.
\newblock Efficient multimodal learning from data-centric perspective.
\newblock \emph{arXiv preprint arXiv:2402.11530}, 2024.

\bibitem[Howard(2017)]{howard2017mobilenets}
Andrew~G Howard.
\newblock Mobilenets: Efficient convolutional neural networks for mobile vision applications.
\newblock \emph{arXiv preprint arXiv:1704.04861}, 2017.

\bibitem[Hudson and Manning(2019)]{hudson2019gqa}
Drew~A Hudson and Christopher~D Manning.
\newblock Gqa: A new dataset for real-world visual reasoning and compositional question answering.
\newblock In \emph{Proceedings of the IEEE/CVF conference on computer vision and pattern recognition}, pages 6700--6709, 2019.

\bibitem[IDEFICS(2023)]{idefics}
IDEFICS.
\newblock Introducing idefics: An open reproduction of state-of-the-art visual language model.
\newblock \url{https://huggingface.co/blog/idefics}, 2023.

\bibitem[Javaheripi et~al.(2023)Javaheripi, Bubeck, Abdin, Aneja, Bubeck, Mendes, Chen, Del~Giorno, Eldan, Gopi, et~al.]{javaheripi2023phi}
Mojan Javaheripi, S{\'e}bastien Bubeck, Marah Abdin, Jyoti Aneja, Sebastien Bubeck, Caio C{\'e}sar~Teodoro Mendes, Weizhu Chen, Allie Del~Giorno, Ronen Eldan, Sivakanth Gopi, et~al.
\newblock Phi-2: The surprising power of small language models.
\newblock \emph{Microsoft Research Blog}, 2023.

\bibitem[Li et~al.(2024{\natexlab{a}})Li, Ge, Ge, Wang, Wang, Zhang, and Shan]{li2024seed}
Bohao Li, Yuying Ge, Yixiao Ge, Guangzhi Wang, Rui Wang, Ruimao Zhang, and Ying Shan.
\newblock Seed-bench: Benchmarking multimodal large language models.
\newblock In \emph{Proceedings of the IEEE/CVF Conference on Computer Vision and Pattern Recognition}, pages 13299--13308, 2024{\natexlab{a}}.

\bibitem[Li et~al.(2024{\natexlab{b}})Li, Wong, Zhang, Usuyama, Liu, Yang, Naumann, Poon, and Gao]{li2024llava}
Chunyuan Li, Cliff Wong, Sheng Zhang, Naoto Usuyama, Haotian Liu, Jianwei Yang, Tristan Naumann, Hoifung Poon, and Jianfeng Gao.
\newblock Llava-med: Training a large language-and-vision assistant for biomedicine in one day.
\newblock \emph{Advances in Neural Information Processing Systems}, 36, 2024{\natexlab{b}}.

\bibitem[Li et~al.(2023{\natexlab{a}})Li, Li, Savarese, and Hoi]{li2023blip}
Junnan Li, Dongxu Li, Silvio Savarese, and Steven Hoi.
\newblock Blip-2: Bootstrapping language-image pre-training with frozen image encoders and large language models.
\newblock \emph{arXiv preprint arXiv:2301.12597}, 2023{\natexlab{a}}.

\bibitem[Li et~al.(2023{\natexlab{b}})Li, Du, Zhou, Wang, Zhao, and Wen]{li2023evaluating}
Yifan Li, Yifan Du, Kun Zhou, Jinpeng Wang, Wayne~Xin Zhao, and Ji-Rong Wen.
\newblock Evaluating object hallucination in large vision-language models.
\newblock \emph{arXiv preprint arXiv:2305.10355}, 2023{\natexlab{b}}.

\bibitem[Li et~al.(2024{\natexlab{c}})Li, Zhang, Wang, Zhong, Chen, Chu, Liu, and Jia]{li2024mini}
Yanwei Li, Yuechen Zhang, Chengyao Wang, Zhisheng Zhong, Yixin Chen, Ruihang Chu, Shaoteng Liu, and Jiaya Jia.
\newblock Mini-gemini: Mining the potential of multi-modality vision language models.
\newblock \emph{arXiv preprint arXiv:2403.18814}, 2024{\natexlab{c}}.

\bibitem[Liang et~al.(2022)Liang, Ge, Tong, Song, Wang, and Xie]{liang2022not}
Youwei Liang, Chongjian Ge, Zhan Tong, Yibing Song, Jue Wang, and Pengtao Xie.
\newblock Not all patches are what you need: Expediting vision transformers via token reorganizations.
\newblock \emph{arXiv preprint arXiv:2202.07800}, 2022.

\bibitem[Lin et~al.(2024)Lin, Tang, Ye, Cui, Zhu, Jin, Zhang, Ning, and Yuan]{lin2024moe}
Bin Lin, Zhenyu Tang, Yang Ye, Jiaxi Cui, Bin Zhu, Peng Jin, Junwu Zhang, Munan Ning, and Li Yuan.
\newblock Moe-llava: Mixture of experts for large vision-language models.
\newblock \emph{arXiv preprint arXiv:2401.15947}, 2024.

\bibitem[Lin et~al.(2014)Lin, Maire, Belongie, Hays, Perona, Ramanan, Doll{\'a}r, and Zitnick]{lin2014microsoft}
Tsung-Yi Lin, Michael Maire, Serge Belongie, James Hays, Pietro Perona, Deva Ramanan, Piotr Doll{\'a}r, and C~Lawrence Zitnick.
\newblock Microsoft coco: Common objects in context.
\newblock In \emph{Computer Vision--ECCV 2014: 13th European Conference, Zurich, Switzerland, September 6-12, 2014, Proceedings, Part V 13}, pages 740--755. Springer, 2014.

\bibitem[Liu et~al.(2024{\natexlab{a}})Liu, Li, Li, and Lee]{liu2024improved}
Haotian Liu, Chunyuan Li, Yuheng Li, and Yong~Jae Lee.
\newblock Improved baselines with visual instruction tuning.
\newblock In \emph{Proceedings of the IEEE/CVF Conference on Computer Vision and Pattern Recognition}, pages 26296--26306, 2024{\natexlab{a}}.

\bibitem[Liu et~al.(2024{\natexlab{b}})Liu, Li, Wu, and Lee]{liu2024visual}
Haotian Liu, Chunyuan Li, Qingyang Wu, and Yong~Jae Lee.
\newblock Visual instruction tuning.
\newblock \emph{Advances in neural information processing systems}, 36, 2024{\natexlab{b}}.

\bibitem[Liu et~al.(2023)Liu, Duan, Zhang, Li, Zhang, Zhao, Yuan, Wang, He, Liu, et~al.]{liu2023mmbench}
Yuan Liu, Haodong Duan, Yuanhan Zhang, Bo Li, Songyang Zhang, Wangbo Zhao, Yike Yuan, Jiaqi Wang, Conghui He, Ziwei Liu, et~al.
\newblock Mmbench: Is your multi-modal model an all-around player?
\newblock \emph{arXiv preprint arXiv:2307.06281}, 2023.

\bibitem[Lloyd(1982)]{lloyd1982least}
Stuart Lloyd.
\newblock Least squares quantization in pcm.
\newblock \emph{IEEE transactions on information theory}, 28\penalty0 (2):\penalty0 129--137, 1982.

\bibitem[Long et~al.(2024)Long, Killick, McCreadie, and Aragon-Camarasa]{long2024robollm}
Zijun Long, George Killick, Richard McCreadie, and Gerardo Aragon-Camarasa.
\newblock Robollm: Robotic vision tasks grounded on multimodal large language models.
\newblock In \emph{2024 IEEE International Conference on Robotics and Automation (ICRA)}, pages 12428--12435. IEEE, 2024.

\bibitem[Lu et~al.(2022)Lu, Mishra, Xia, Qiu, Chang, Zhu, Tafjord, Clark, and Kalyan]{lu2022learn}
Pan Lu, Swaroop Mishra, Tanglin Xia, Liang Qiu, Kai-Wei Chang, Song-Chun Zhu, Oyvind Tafjord, Peter Clark, and Ashwin Kalyan.
\newblock Learn to explain: Multimodal reasoning via thought chains for science question answering.
\newblock \emph{Advances in Neural Information Processing Systems}, 35:\penalty0 2507--2521, 2022.

\bibitem[MacQueen(1967)]{macqueen1967some}
J MacQueen.
\newblock Some methods for classification and analysis of multivariate observations.
\newblock In \emph{Proceedings of 5-th Berkeley Symposium on Mathematical Statistics and Probability/University of California Press}, 1967.

\bibitem[Meng et~al.(2024)Meng, Yang, Tian, Dai, Wu, Gao, and Jiang]{meng2024deepstack}
Lingchen Meng, Jianwei Yang, Rui Tian, Xiyang Dai, Zuxuan Wu, Jianfeng Gao, and Yu-Gang Jiang.
\newblock Deepstack: Deeply stacking visual tokens is surprisingly simple and effective for lmms.
\newblock \emph{arXiv preprint arXiv:2406.04334}, 2024.

\bibitem[Mesk{\'o}(2023)]{mesko2023impact}
Bertalan Mesk{\'o}.
\newblock The impact of multimodal large language models on health care’s future.
\newblock \emph{Journal of medical Internet research}, 25:\penalty0 e52865, 2023.

\bibitem[Niu et~al.(2024)Niu, Sharma, Biamby, Quenum, Bai, Shi, Darrell, and Herzig]{niu2024llarva}
Dantong Niu, Yuvan Sharma, Giscard Biamby, Jerome Quenum, Yutong Bai, Baifeng Shi, Trevor Darrell, and Roei Herzig.
\newblock Llarva: Vision-action instruction tuning enhances robot learning.
\newblock \emph{arXiv preprint arXiv:2406.11815}, 2024.

\bibitem[Radford et~al.(2021)Radford, Kim, Hallacy, Ramesh, Goh, Agarwal, Sastry, Askell, Mishkin, Clark, et~al.]{radford2021learning}
Alec Radford, Jong~Wook Kim, Chris Hallacy, Aditya Ramesh, Gabriel Goh, Sandhini Agarwal, Girish Sastry, Amanda Askell, Pamela Mishkin, Jack Clark, et~al.
\newblock Learning transferable visual models from natural language supervision.
\newblock In \emph{International conference on machine learning}, pages 8748--8763. PMLR, 2021.

\bibitem[Radosavovic et~al.(2020)Radosavovic, Kosaraju, Girshick, He, and Doll{\'a}r]{radosavovic2020designing}
Ilija Radosavovic, Raj~Prateek Kosaraju, Ross Girshick, Kaiming He, and Piotr Doll{\'a}r.
\newblock Designing network design spaces.
\newblock In \emph{Proceedings of the IEEE/CVF conference on computer vision and pattern recognition}, pages 10428--10436, 2020.

\bibitem[Sedgwick(2014)]{sedgwick2014spearman}
Philip Sedgwick.
\newblock Spearman's rank correlation coefficient.
\newblock \emph{Bmj}, 349, 2014.

\bibitem[Shang et~al.(2024)Shang, Cai, Xu, Lee, and Yan]{shang2024llava}
Yuzhang Shang, Mu Cai, Bingxin Xu, Yong~Jae Lee, and Yan Yan.
\newblock Llava-prumerge: Adaptive token reduction for efficient large multimodal models.
\newblock \emph{arXiv preprint arXiv:2403.15388}, 2024.

\bibitem[Singh et~al.(2019)Singh, Natarajan, Shah, Jiang, Chen, Batra, Parikh, and Rohrbach]{singh2019towards}
Amanpreet Singh, Vivek Natarajan, Meet Shah, Yu Jiang, Xinlei Chen, Dhruv Batra, Devi Parikh, and Marcus Rohrbach.
\newblock Towards vqa models that can read.
\newblock In \emph{Proceedings of the IEEE/CVF conference on computer vision and pattern recognition}, pages 8317--8326, 2019.

\bibitem[Team et~al.(2023)Team, Anil, Borgeaud, Wu, Alayrac, Yu, Soricut, Schalkwyk, Dai, Hauth, et~al.]{team2023gemini}
Gemini Team, Rohan Anil, Sebastian Borgeaud, Yonghui Wu, Jean-Baptiste Alayrac, Jiahui Yu, Radu Soricut, Johan Schalkwyk, Andrew~M Dai, Anja Hauth, et~al.
\newblock Gemini: a family of highly capable multimodal models.
\newblock \emph{arXiv preprint arXiv:2312.11805}, 2023.

\bibitem[Touvron et~al.(2023{\natexlab{a}})Touvron, Lavril, Izacard, Martinet, Lachaux, Lacroix, Rozi{\`e}re, Goyal, Hambro, Azhar, et~al.]{touvron2023llama}
Hugo Touvron, Thibaut Lavril, Gautier Izacard, Xavier Martinet, Marie-Anne Lachaux, Timoth{\'e}e Lacroix, Baptiste Rozi{\`e}re, Naman Goyal, Eric Hambro, Faisal Azhar, et~al.
\newblock Llama: Open and efficient foundation language models.
\newblock \emph{arXiv preprint arXiv:2302.13971}, 2023{\natexlab{a}}.

\bibitem[Touvron et~al.(2023{\natexlab{b}})Touvron, Martin, Stone, Albert, Almahairi, Babaei, Bashlykov, Batra, Bhargava, Bhosale, et~al.]{touvron2023llama2}
Hugo Touvron, Louis Martin, Kevin Stone, Peter Albert, Amjad Almahairi, Yasmine Babaei, Nikolay Bashlykov, Soumya Batra, Prajjwal Bhargava, Shruti Bhosale, et~al.
\newblock Llama 2: Open foundation and fine-tuned chat models.
\newblock \emph{arXiv preprint arXiv:2307.09288}, 2023{\natexlab{b}}.

\bibitem[Walmer et~al.(2023)Walmer, Suri, Gupta, and Shrivastava]{walmer2023teaching}
Matthew Walmer, Saksham Suri, Kamal Gupta, and Abhinav Shrivastava.
\newblock Teaching matters: Investigating the role of supervision in vision transformers.
\newblock In \emph{Proceedings of the IEEE/CVF Conference on Computer Vision and Pattern Recognition}, pages 7486--7496, 2023.

\bibitem[Wang et~al.(2023{\natexlab{a}})Wang, Chen, Lin, Ding, Liu, Bao, Yan, and Ding]{wang2023hierarchical}
Ao Wang, Hui Chen, Zijia Lin, Zixuan Ding, Pengzhang Liu, Yongjun Bao, Weipeng Yan, and Guiguang Ding.
\newblock Hierarchical prompt learning using clip for multi-label classification with single positive labels.
\newblock In \emph{Proceedings of the 31st ACM International Conference on Multimedia}, pages 5594--5604, 2023{\natexlab{a}}.

\bibitem[Wang et~al.(2023{\natexlab{b}})Wang, Chen, Lin, Zhao, Han, and Ding]{wang2023cait}
Ao Wang, Hui Chen, Zijia Lin, Sicheng Zhao, Jungong Han, and Guiguang Ding.
\newblock Cait: Triple-win compression towards high accuracy, fast inference, and favorable transferability for vits.
\newblock \emph{arXiv preprint arXiv:2309.15755}, 2023{\natexlab{b}}.

\bibitem[Wang et~al.(2024{\natexlab{a}})Wang, Chen, Lin, Han, and Ding]{wang2024repvit}
Ao Wang, Hui Chen, Zijia Lin, Jungong Han, and Guiguang Ding.
\newblock Repvit: Revisiting mobile cnn from vit perspective.
\newblock In \emph{Proceedings of the IEEE/CVF Conference on Computer Vision and Pattern Recognition}, pages 15909--15920, 2024{\natexlab{a}}.

\bibitem[Wang et~al.(2024{\natexlab{b}})Wang, Chen, Liu, Chen, Lin, Han, and Ding]{wang2024yolov10}
Ao Wang, Hui Chen, Lihao Liu, Kai Chen, Zijia Lin, Jungong Han, and Guiguang Ding.
\newblock Yolov10: Real-time end-to-end object detection.
\newblock \emph{arXiv preprint arXiv:2405.14458}, 2024{\natexlab{b}}.

\bibitem[Wang et~al.(2024{\natexlab{c}})Wang, Wu, Li, Jiang, Shu, Shi, Hu, Ma, Liu, Wang, et~al.]{wang2024large}
Jiaqi Wang, Zihao Wu, Yiwei Li, Hanqi Jiang, Peng Shu, Enze Shi, Huawen Hu, Chong Ma, Yiheng Liu, Xuhui Wang, et~al.
\newblock Large language models for robotics: Opportunities, challenges, and perspectives.
\newblock \emph{arXiv preprint arXiv:2401.04334}, 2024{\natexlab{c}}.

\bibitem[Wang et~al.(2023{\natexlab{c}})Wang, Lv, Yu, Hong, Qi, Wang, Ji, Yang, Zhao, Song, et~al.]{wang2023cogvlm}
Weihan Wang, Qingsong Lv, Wenmeng Yu, Wenyi Hong, Ji Qi, Yan Wang, Junhui Ji, Zhuoyi Yang, Lei Zhao, Xixuan Song, et~al.
\newblock Cogvlm: Visual expert for pretrained language models.
\newblock \emph{arXiv preprint arXiv:2311.03079}, 2023{\natexlab{c}}.

\bibitem[Wang et~al.(2023{\natexlab{d}})Wang, Xie, Hu, Zou, Fan, Tong, Wen, Wu, Deng, Li, et~al.]{wang2023drivemlm}
Wenhai Wang, Jiangwei Xie, ChuanYang Hu, Haoming Zou, Jianan Fan, Wenwen Tong, Yang Wen, Silei Wu, Hanming Deng, Zhiqi Li, et~al.
\newblock Drivemlm: Aligning multi-modal large language models with behavioral planning states for autonomous driving.
\newblock \emph{arXiv preprint arXiv:2312.09245}, 2023{\natexlab{d}}.

\bibitem[Xiao et~al.(2024)Xiao, Zhou, Liu, Liu, Li, Liu, and Huang]{xiao2024comprehensive}
Hanguang Xiao, Feizhong Zhou, Xingyue Liu, Tianqi Liu, Zhipeng Li, Xin Liu, and Xiaoxuan Huang.
\newblock A comprehensive survey of large language models and multimodal large language models in medicine.
\newblock \emph{arXiv preprint arXiv:2405.08603}, 2024.

\bibitem[Xu et~al.(2024)Xu, Zhang, Xie, Zhao, Guo, Wong, Li, and Zhao]{xu2024drivegpt4}
Zhenhua Xu, Yujia Zhang, Enze Xie, Zhen Zhao, Yong Guo, Kwan-Yee~K Wong, Zhenguo Li, and Hengshuang Zhao.
\newblock Drivegpt4: Interpretable end-to-end autonomous driving via large language model.
\newblock \emph{IEEE Robotics and Automation Letters}, 2024.

\bibitem[Yao et~al.(2024)Yao, Li, Ren, Wang, Liu, Sun, and Hou]{yao2024deco}
Linli Yao, Lei Li, Shuhuai Ren, Lean Wang, Yuanxin Liu, Xu Sun, and Lu Hou.
\newblock Deco: Decoupling token compression from semantic abstraction in multimodal large language models.
\newblock \emph{arXiv preprint arXiv:2405.20985}, 2024.

\bibitem[Ye et~al.(2024)Ye, Gan, Huang, Ge, Shan, and Tang]{ye2024voco}
Xubing Ye, Yukang Gan, Xiaoke Huang, Yixiao Ge, Ying Shan, and Yansong Tang.
\newblock Voco-llama: Towards vision compression with large language models.
\newblock \emph{arXiv preprint arXiv:2406.12275}, 2024.

\bibitem[Yu et~al.(2023)Yu, Yang, Li, Wang, Lin, Liu, Wang, and Wang]{yu2023mm}
Weihao Yu, Zhengyuan Yang, Linjie Li, Jianfeng Wang, Kevin Lin, Zicheng Liu, Xinchao Wang, and Lijuan Wang.
\newblock Mm-vet: Evaluating large multimodal models for integrated capabilities.
\newblock \emph{arXiv preprint arXiv:2308.02490}, 2023.

\bibitem[Yuan et~al.(2023)Yuan, Li, and Sun]{yuan2023tinygpt}
Zhengqing Yuan, Zhaoxu Li, and Lichao Sun.
\newblock Tinygpt-v: Efficient multimodal large language model via small backbones.
\newblock \emph{arXiv preprint arXiv:2312.16862}, 2023.

\bibitem[Yue et~al.(2024)Yue, Chen, Geiping, Li, Goldstein, and Lim]{yue2024object}
Kaiyu Yue, Bor-Chun Chen, Jonas Geiping, Hengduo Li, Tom Goldstein, and Ser-Nam Lim.
\newblock Object recognition as next token prediction.
\newblock In \emph{Proceedings of the IEEE/CVF Conference on Computer Vision and Pattern Recognition}, pages 16645--16656, 2024.

\bibitem[Zheng et~al.(2023)Zheng, Chiang, Sheng, Zhuang, Wu, Zhuang, Lin, Li, Li, Xing, et~al.]{zheng2023judging}
Lianmin Zheng, Wei-Lin Chiang, Ying Sheng, Siyuan Zhuang, Zhanghao Wu, Yonghao Zhuang, Zi Lin, Zhuohan Li, Dacheng Li, Eric Xing, et~al.
\newblock Judging llm-as-a-judge with mt-bench and chatbot arena.
\newblock \emph{Advances in Neural Information Processing Systems}, 36:\penalty0 46595--46623, 2023.

\bibitem[Zhu et~al.(2023)Zhu, Chen, Shen, Li, and Elhoseiny]{zhu2023minigpt}
Deyao Zhu, Jun Chen, Xiaoqian Shen, Xiang Li, and Mohamed Elhoseiny.
\newblock Minigpt-4: Enhancing vision-language understanding with advanced large language models.
\newblock \emph{arXiv preprint arXiv:2304.10592}, 2023.

\bibitem[Zhu et~al.(2024)Zhu, Zhu, Liu, Ou, Mou, and Tang]{zhu2024llavaphi}
Yichen Zhu, Minjie Zhu, Ning Liu, Zhicai Ou, Xiaofeng Mou, and Jian Tang.
\newblock Llava-phi: Efficient multi-modal assistant with small language model, 2024.

\end{thebibliography}
}

\end{document}